\pdfoutput=1

\documentclass[11pt,usenames,dvipsnames]{article}

\usepackage{eacl2024}

\usepackage{times}
\usepackage{latexsym}

\usepackage[T1]{fontenc}

\usepackage[utf8]{inputenc}

\usepackage{microtype}

\usepackage{inconsolata}

\usepackage{comment}
\usepackage{adjustbox}
\usepackage{multirow}
\usepackage{booktabs}
\usepackage{csquotes}
\usepackage{amsmath}
\usepackage{graphicx}
\graphicspath{ {./images/} }
\title{Large Language Models for Scientific Information Extraction: \\ An Empirical Study for Virology}

\author{Mahsa Shamsabadi \and Jennifer D'Souza \and Sören Auer \\
         TIB Leibniz Information Centre for Science and Technology, \\ Hannover, Germany \\
        \texttt{\{mahsa.shamsabasdi,jennifer.dsouza,auer\}@tib.eu}}

\begin{document}
\maketitle
\begin{abstract}









In this paper, we champion the use of structured and semantic content representation of discourse-based scholarly communication, inspired by tools like Wikipedia infoboxes or structured Amazon product descriptions. These representations provide users with a concise overview, aiding scientists in navigating the dense academic landscape. Our novel automated approach leverages the robust text generation capabilities of LLMs to produce structured scholarly contribution summaries, offering both a practical solution and insights into LLMs' emergent abilities.


For LLMs, the prime focus is on improving their general intelligence as conversational agents. We argue that these models can also be applied effectively in information extraction (IE), specifically in complex IE tasks within terse domains like Science. This paradigm shift replaces the traditional modular, pipelined machine learning approach with a simpler objective expressed through instructions. Our results show that finetuned FLAN-T5 with 1000x fewer parameters than the state-of-the-art GPT-davinci is competitive for the task.


\end{abstract}

\section{Introduction}

\begin{figure}[!tb]
\includegraphics[width=\linewidth]{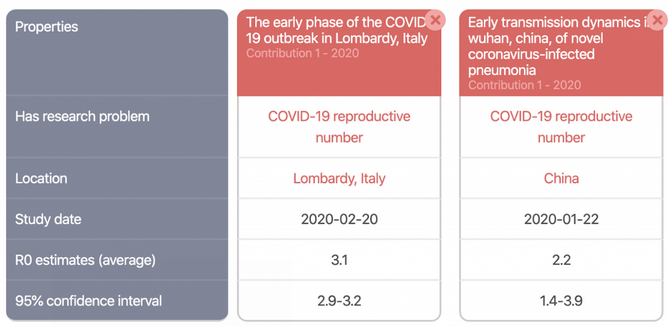}
\caption{Two structured research contributions compared in the Open Research Knowledge Graph (papers in columns, properties in rows and values in cells).}
\label{orkg-comp}
\end{figure}

Scholarly communication in the digital age is facing significant challenges due to the overwhelming volume of publications~\cite{johnson2018stm} thereby creating the need for efficient access to relevant knowledge. In this regard, next-generation scholarly digital libraries, such as the Open Research Knowledge Graph (\href{https://orkg.org/}{ORKG})~\cite{auer2020improving,stocker2023fair}, offer a promising solution by adopting semantic publishing principles~\cite{shotton2009semantic}. The ORKG stores \textit{scholarly contributions} in a structured and semantic way, leveraging a knowledge graph (KG) representation~\cite{ehrlinger2016towards,fensel2020introduction}. The fine-grained semantic contribution representation in the ORKG utilizes property-value tuples, capturing important aspects and corresponding observations of research contributions. This representation enhances understanding and navigation of scholarly content by both humans and machines. With selected properties that apply universally to research on a specific problem, the ORKG enables intelligent exploration and assistance services, including \href{https://orkg.org/comparisons}{research comparisons} based on shared properties, e.g., \autoref{orkg-comp}. Its novel information access methods provide condensed overviews of the state-of-the-art, supporting strategic reading~\cite{renear2009strategic} in the ever-growing publication landscape. 

\begin{figure*}[!t]
\includegraphics[width=\textwidth]{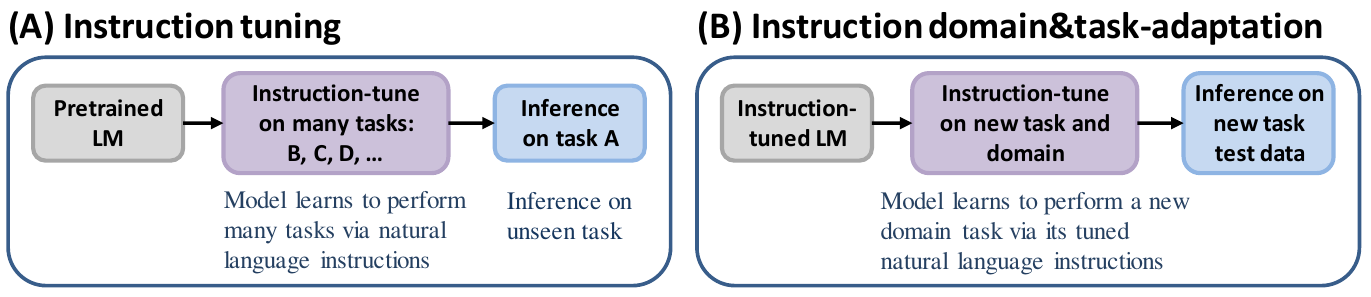}
\caption{Comparing (A) instruction tuning with (B) instruction-tuned LLM domain- and task-tuning of this work.}
\label{instruction-tuning}
\end{figure*}

This work, as a text mining service toward producing scalable solutions for the ORKG, for the first time, introduces a complex information extraction (IE) task. Our notion of complex IE entails joint entity and relation extraction in a single objective aligned with the structured property-value format of contributions in the ORKG. We defined the complex IE task w.r.t. a key research problem in the domain of Epidemics \& Virology, i.e. estimating the basic reproduction number ($R0$) for infectious diseases. This $R0$ estimate research topic was brought to common knowledge during the recent Covid-19 pandemic by the Centers for Disease Control and Prevention (\href{https://www.cdc.gov/}{CDC}) as a \href{https://www.cdc.gov/coronavirus/2019-ncov/hcp/planning-scenarios.html}{key informant}. Important to infectious disease epidemiology, generally, the $R0$ estimate represents the average number of secondary infections caused by a single infected individual~\cite{r0-def}. In other words, it is an estimate of disease progression in a given population. E.g., the estimated R0 for COVID-19 has been reported between 2.5 to 5.7~\cite{sanche2020novel}. It varies for different infectious diseases and populations. For researchers in Epidemics \& Virology, it is interesting to be able to compare the R0 of different viruses facilitated by structured contribution data available in the ORKG. The alternative, traditional, and seemingly impossible knowledge comprehension task, would be to scour for vital information buried in unstructured text across the 44k articles by \href{https://scholar.google.com/scholar?hl=en&as_sdt=0%2C5&q=COVID-19+R0&btnG=}{Covid-19 R0 estimate} Google search.

To define our complex IE task, an expert semantic modeler created a \href{https://orkg.org/comparison/R44930/}{research comparison} based on structured property-value pairs for Covid-19 $R0$ estimate contributions across 30 abstracts. Consequently, six properties were modeled: \textit{disease name}, \textit{location}, \textit{date}, \textit{R0 value}, \textit{\%CI values},\footnote{CI stands for confidence interval.} and \textit{method}. The semantic modeling aimed to identify properties that were both generic enough to structure most related research on the $R0$ estimate (in the context of a research comparison) and specialized enough to reflect the vital details of the $R0$ contribution (by identifying commonalities in observations reported across 30 different abstracts). 
This structured format is called \textsc{orkg-R0}. Thus our complex IE task focused on extracting property-value pairs for \textsc{orkg-R0} contributions in scholarly article abstracts. To address this task, a larger gold-standard corpus was annotated (details in \autoref{corpus}) and an LLM-based solution was optimally designed (introduced next, details in \autoref{model}).


\begin{figure*}[!tb]
\includegraphics[width=\textwidth]{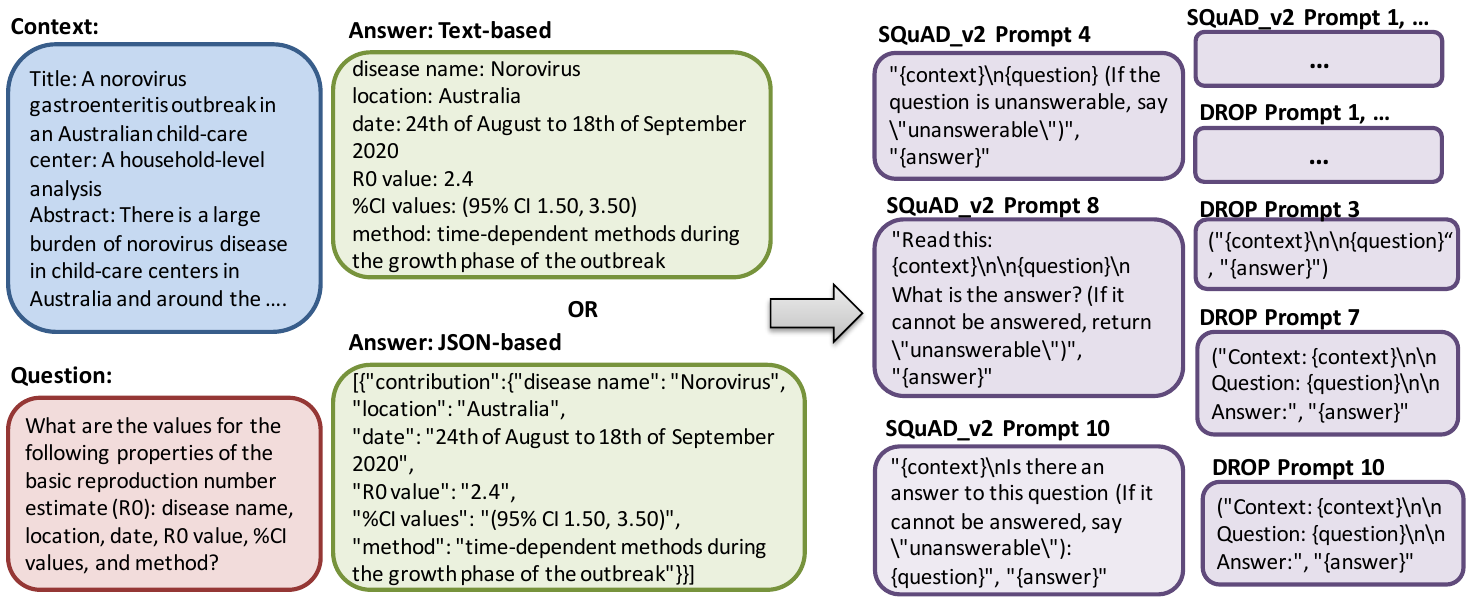}
\caption{Multiple instruction prompts describing our complex scientific information extraction (IE) task.}
\label{learning-model}
\end{figure*}

The complex IE task introduced earlier is addressed as single-task instruction-based finetuning of an instruction-tuned Large Language Model (LLM) with the primary objective of \textit{better aligning the LLM to our task and domain}. Our approach is characterized in \autoref{instruction-tuning}. We chose LLMs for their rich parameter spaces and ability to handle complex IE tasks with simple instruction prompts~\cite{instructgpt}. Unlike traditional pipelined-based IE, which are prone to error propagation and require extensive manual engineering, LLMs offer flexibility, adaptability, and the ability to handle a wide range of tasks in zero- and few-shot settings through instructions~\cite{gpt2,gpt3,flan}. By relying on instruction prompting, we can effectively address complex inter-relations without the need for an exhaustive enumeration of all possible relations or preliminary named entity recognition (NER). We finetune an LLM from the sequence-to-sequence encoder-decoder-based T5 model class~\cite{t5} to accept a research paper title and abstract and instruct it to write the \textsc{orkg-R0} structured ``summary'' of knowledge in the prompt as either text-based or as a structured JSON object. For the LLM, we specifically select the instruction-tuned FLAN-T5-Large model~\cite{flan-t5} with reported 780M parameters. There could have been one of two directions for this work: scaling the models or instruction fine-tuning of a moderate-sized LLM, i.e. with parameters in millions versus 1000x more in billions. We chose the latter. We believe that our choice makes model tuning more accessible within the research community while empirically proving to be nonetheless effective (experimental details in \autoref{eval}). Furthermore, our choice of Google's FLAN-T5, open-sourced and easily accessible in the \href{https://huggingface.co/docs/transformers/model_doc/flan-t5}{Transformers library}, obviates any paywall that hinders access to LLMs for the research community at large. 
For instruction-based finetuning, we use applicable instructions from the \textit{open-sourced instruction generalization efforts} introduced as the ``Flan 2022 Collection''~\citep{flan-collection}. Our approach differs from finetuning a pretrained LM as we instead finetune an instruction-tuned LM, enabling the model to effectively follow instructions it has been trained on and adapt to a new domain and complex IE task, without the need to handle variability in learning new instruction formats. Our approach is shown in \autoref{learning-model}.


In this context, the central research question (\textbf{RQ}) of this work examines: \textit{How does instruction-based finetuning enhance LLM performance in a unique domain, specifically in a complex scientific field like Virology that requires specialized expertise?} Summarily, the main contributions of our work are as follows: 1) \textbf{Corpus}: A \href{https://doi.org/10.5281/zenodo.8068441}{gold-standard corpus} of 1,500 annotated structured abstracts based on \textsc{orkg-R0}. 2) \textbf{Methodological}: We adopt ``single-task instruction-finetuning'' to enhance LLMs' domain and task adaptation. It involves selecting instructions from the open-sourced FLAN collection and fine-tuning FLAN-T5 780M to respond to those instructions. Our \href{https://anonymous.4open.science/r/R0_Structured_Information_Extraction-5920/README.md}{source code} is released. 3) \textbf{Methodological}: Our approach distinguishes itself in the realm of IE research by introducing an LLM-based approach that breaks away from traditional pipeline-based methods for entity and relation extraction. Instead, we propose a single-system approach utilizing a moderately-sized LLM, which holds potential for practical applications. And 4) \textbf{Results}: Our instruction-finetuned \textsc{orkg}-FLAN-T5$_{R0}$ 780M outperforms pretrained T5, instruction-tuned FLAN-T5, and GPT3.5-davinci 175B on \textsc{orkg-R0} complex IE.




\section{Background: Scholarly Communication}

Semantic scholarly knowledge publishing models, such as the ORKG, specifically the \textsc{orkg-R0} instance in this work, and the structured abstracts methodology (e.g., \href{https://www.nlm.nih.gov/bsd/policy/structured_abstracts.html}{IMRAD}) employed by publishers like \href{https://pubmed.ncbi.nlm.nih.gov/}{PubMed} have distinct approaches and serve different purposes in scholarly communication. This section distinguishes the two.

The ORKG~\cite{auer_soren_2018} and similar semantic knowledge publishing models~\cite{baas2020scopus,birkle2020web,wang2020microsoft,aryani2018research,manghi_paolo_2019_3516918,hendricks2020crossref,fricke2018semantic} aim to create interconnected and machine-actionable representations of scholarly knowledge. They leverage semantic technologies, knowledge graphs (KGs), and ontologies to capture the meaning, context, and relationships between research concepts. The ORKG, for example, stores scholarly contributions as structured property-value pairs, enabling advanced exploration, comparison~\cite{oelen2019comparing}, and analysis via visualizations~\cite{wiens2020towards} of research findings. The strength of semantic knowledge publishing models lies in their ability to facilitate interdisciplinary collaborations, data integration, and automated processing of scholarly information. They enhance research transparency, enable advanced search and discovery, and support the development of novel strategic reading tools and services for researchers.

On the other hand, the structured abstracts methodology~\cite{haynes1990more,hayward1993more,nakayama2005adoption,kulkarni1996structured,hopewell2008consort}, e.g., IMRAD~\cite{sollaci2004introduction}, focuses on organizing research articles into a specific format. IMRAD advocates for a structured abstract based on four points, viz. Introduction, Methods, Results, and Discussion, to provide a standardized framework for reporting research. The strength of structured abstracts lies in their ability to provide a clear and consistent organization of research findings. They help readers quickly understand the key components of a study and locate specific information within the article. Structured abstracts facilitate efficient scanning and information retrieval.


In summary, semantic scholarly knowledge publishing models enhance the machine-actionability and interoperability of scholarly knowledge, enabling advanced computational exploration and analysis. They offer opportunities for interdisciplinary collaborations and innovative research tools. On the other hand, structured abstracts provide a standardized format for reporting research, facilitating efficient information retrieval.

\section{Corpus}
\label{corpus}
We aim to create a high-quality corpus for the complex scientific IE task introduced in this work. The corpus creation goal was to obtain gold-standard property-value structured format representation w.r.t. the six predicates in \textsc{orkg-R0} from scholarly article abstracts. These structured representations encapsulate the R0 estimate research problem for infectious diseases.

\noindent{\textbf{Base corpus.}} Our starting point was the large-scale \href{https://github.com/allenai/cord19}{CORD-19} dataset~\citep{wang2020cord} provided by AllenAI. 
This resource comprised a growing collection of publications and preprints on Covid-19, its variants, related historical coronaviruses such as SARS and MERS, as well as other infectious diseases such as H1N1 Influenza, Dengue, Monkeypox, Ebola, Zika virus, Norovirus, etc. At our download date timestamp 2022-06-02 it comprised over 800,000 total publications. The dataset covered diverse topics such as epidemiology, virology, clinical studies, public health, and more. It served as a valuable resource for researchers, policymakers, and the general public to access and analyze the latest scientific knowledge related to COVID-19. Since CORD-19 contained articles on various themes, as a next step the corpus was filtered to include only articles on the $R0$ estimate theme.

\noindent{\textbf{Corpus filtering.}} Our method for filtering the base corpus to our desired collection was simple. We implemented \href{https://anonymous.4open.science/r/R0_Structured_Information_Extraction-5920/src/data/cord_extraction_and_processing/extract_data_from_cord_metadata.py}{pattern-based heuristics} using variants of the phrase ``R0 estimate'' and checked the publication abstract for containment. The base corpus was then reduced to  4590 instances. Post deduplication, the collection was further reduced to 3967 instances. Other than exact duplicates, there were other near-duplication patterns such as punctuation marks stripped or retained, numbers with or without decimal points saved as different data instances. Near-duplicates were also filtered by clustering abstracts that were 95\% similar (583 clusters from 1227 articles were created). A human annotator went through all clusters and decided on one abstract to retain while dropping all others. The resulting curated corpus contained 3024 abstracts which included a direct mention or a variant of the phrase ``R0 estimate''.

\noindent{\textbf{The \textsc{orkg-R0} model.}} Here we provide an explanation of \textsc{orkg-R0} as an ideal representation of a structured contribution for the research problem of ``$R0$ estimate,'' as defined by an expert semantic modeler. The $R0$ estimate pertains to an infectious disease (\textit{disease name}), for a specific population demographic (\textit{location}), with validity for a specific time period (\textit{date}). It reports a specific value (\textit{$R0$ value}), along with a confidence interval for the statistical value (\textit{\%CI values}), and is computed by a statistical method (\textit{method}). 

\noindent{\textbf{Annotation exercise.}} To ensure a practical and realistic human annotation target, we selected a sub-sample of 1500 articles from the curated 3024 dataset. This would then serve as the gold-standard dataset for training and development purposes, as an empirical basis for future research. A team of two annotators produced the \textsc{orkg-R0} structured annotations with the corpus raw data comprising a paper title and abstract, where each instance is uniquely identified by a cord\_id. The overall annotation exercise lasted 3 months. The annotation task began by distinguishing between the papers actually reporting an $R0$ value as a contribution and those that just mentioned the ``R0 estimate'' keyword in the abstract, but did not actually report a value as a contribution of the work. Resultingly, we found 652 articles reported an $R0$ value and thus were annotated for the \textsc{orkg-R0} structure (referred to as the ``answerable'' set, in short \textit{ans}), while 850 did not (referred to as the ``unanswerable'' set, in short \textit{unans}). Among the 652 articles, approximately 157 had multiple contributions for the ``$R0$ estimate.'' Notably, a few articles stood out with 10, 11, or 16 reported contributions. The gold-standard annotated set was made available in two formats: text-based and JSON-based, which are illustrated by the green boxes in \autoref{learning-model}. In the text-based format, multiple contributions were separated using a pipe character, while in the JSON format, they were encoded as separate JSON object dictionaries. We observed that the JSON data structure is more conducive for utilization in downstream applications. Therefore, our empirical analysis regarding LLMs aimed not only to assess their ability to generate structured abstract summaries but also to evaluate their compatibility with a specific data structure. This allows for the seamless integration of their output into downstream applications.

\noindent{\textbf{The annotators.}} In our annotation process, we first developed a structured summary model for the ``R0 estimate for infectious diseases'' using both domain experts and a semantic modeler specializing in ontology design. Next, a PhD student populated the model using a dataset of abstracts, treating it as a form-filling task of reported facts. While the task itself is tedious in that the student needed to read all abstracts to populate the properties, the process did not entail much ambiguity in the decisions. The definition of the properties we selected are fairly straightforward and the values are to be directly extracted from the text. For discrepancies in spans for the values selected, the LLM is expected to be robust enough to arrive at the optimal extraction scenario. For any concerns on quality, our gold-standard test dataset annotations versus the LLM predictions eventually obtained can be publicly browsed at this link \url{https://scinext-project.github.io/#/r0-estimates}.

\noindent{\textbf{Our complex IE task objective.}} We phrased the following question to formulate our task objective w.r.t. the \textsc{orkg-R0} extraction target: \textit{What are the values for the following properties of the basic reproduction number estimate (R0): disease name, location, date, R0 value, \%CI values, and method?} In essence, it encapsulates an IE task. 

\noindent{\textbf{Instructions for the LLM.}} Instruction tuning is a novel approach~\cite{unifiedqa,mccann2018natural,keskar2019unifying} that improves LLMs' performance by providing explicit instructions during finetuning, guiding the model's behavior~\cite{instructgpt,flan-t5,metaicl} and enhancing its adaptability and effectiveness in diverse learning scenarios. Unlike traditional non-instruction tuning methods~\cite{t5,liu2019multi,muppet,ext5} that rely solely on unlabeled data, instruction tuning incorporates specific guidance, simplifying the finetuning process and enabling better performance on new tasks and domains~\cite{sanh2022multitask}. It became possible to generically prompt an LLM to perform different tasks with a single instruction. As such it can be considered as a template that encodes the task and its objective, in turn telling the LLM what to do with the given objective.

The ``Flan 2022 Collection'' was a large-scale open-sourced collection of 62 prior publicly released datasets in the NLP community clustered as 12 task types, such as reading comprehension (RC), sentiment, natural language inference (NLI), struct to text, etc. It is the most comprehensive resource facilitating open-sourced LLM development as generic multi-task models. Importantly, and of relevance to this work, FLAN was not just a super-amalgamation of datasets encapsulating different learning objectives, but also included at least \href{https://github.com/google-research/FLAN/blob/main/flan/templates.py}{10 human-curated natural instructions} per dataset that described the task for that dataset. As such, we select a set of instructions to guide the LLM for our complex IE task from the FLAN collection. Specifically, we identified the \href{https://anonymous.4open.science/r/R0_Structured_Information_Extraction-5920/src/data/create_templated_datasets/build_templated_datasets.py}{applicable instructions} to our task were those designed for the SQuAD\_v2~\cite{squad-v1,squad-v2} and DROP~\cite{drop} datasets. The general characteristic of the selected instructions is that they encode a context (in our case the paper title and abstract) and the task objective, and instruct the model to fulfill the objective. The purple boxes in \autoref{learning-model} show some exemplars. Examples of all instructions are in \autoref{app:temp}.



Our work is positioned here, coalescing the most relevant collection of instructions that were used to instruction-finetune the T5~\citeyearpar{t5} model class, as the strong reference point for any future open source work on single-task instruction finetuning.

\section{Approach}
\label{model}


Our approach is single-task instruction-finetuning for our novel introduced complex IE task. As such it aims to be an incremental progression of the instruction-tuning paradigm introduced as FLAN (Finetuned Language Net)~\citeyearpar{flan,flan-t5,flan-collection}. Specifically \citet{flan-t5} ask: \textit{are instruction-finetuned models better for single-task finetuning?} as a recommendation for future work. Our work then is a direct examination of this research question except for a novel task type that we also introduce for the first time in the community. 


Now, we outline our methodology. \textbf{Step 1.} Collect relevant instructions for \textsc{orkg-R0} complex IE to guide an LLM towards the desired objective. \textbf{Step 2.} Instantiate the instructions to the LLM using gold-standard structured data and a formulated question (e.g. in \autoref{app:temp}). \textbf{Step 3.} Finetune the LLM with the instruction-instantiated data. Three training strategies are explored: single-instruction tuning, all-instruction tuning, and best-instruction tuning based on evaluation results.


\subsection{Model}


We adopt the FLAN-T5 model~\cite{flan-t5} wr.r.t. its \href{https://github.com/google-research/t5x/blob/main/docs/models.md#flan-t5-checkpoints}{public checkpoints}. This encoder-decoder sequence generation model is available in a range of sizes: Small 80M, Base 250M, Large 780M, XL 3B, and XXL 11B. We choose the Large model as a middle ground between the Small and XXL models, providing enough parameters for our complex IE task and practicality for deployment. Additionally, we find it inefficient to test extreme scale LLMs for a single task. Our choice of Flan-T5 was motivated by prior empiricism~\cite{flan-collection} proving instruction-tuned models as more computationally efficient starting checkpoints for new tasks -- FLAN-T5 required less finetuning to converge higher and faster than T5 on single downstream tasks~\citeyearpar{flan-collection}. Our model choice builds upon previous research, enhancing the T5 text-to-text sequence generation model~\citeyearpar{t5} with FLAN-T5~\citeyearpar{flan-t5} to improve alignment with instructions in unseen tasks and zero-shot settings. Our resulting model is called \textsc{orkg}-FLAN-T5$_{R0}$.







\begin{table*}[!htb]
  \centering
  \begin{adjustbox}{width=1\textwidth}
  
    \begin{tabular}{|p{2cm}|p{1.1cm}|rrrrr|rrrrr|}
      \hline
      & & \multicolumn{5}{|c|}{Highest Scores} & \multicolumn{5}{c|}{Lowest Scores} \\
      \toprule
    \bf Model & \bf Format & \bf Rouge1 & \bf Rouge2 & \bf RougeL & \bf RougeLsum & \bf \stackbox[c]{General\\-Accuracy} &\
       \bf  Rouge1 & \bf Rouge2 & \bf RougeL & \bf RougeLsum & \bf \stackbox[c]{General\\-Accuracy}\\
      \midrule
\multirow{2}{*}{T5} & text & 12.46 &	4.56 &	10.37 &	11.99 &	45.00 & 1.37 &	0.52 &	1.21 &	1.37 &	45.00 \\ \cline{2-12}
     & json & 12.01 &	4.33 &	10.54 &	10.49 &	45.00 & 1.35 &	0.51 &	1.18 &	1.17 &	45.00 \\ \hline   
 
\multirow{2}{*}{FLAN-T5} & text & 51.66 & 0.42 & 51.42 & 51.85 & 56.33 & 7.94 & 3.98 & 7.68 & 7.85 & 45.00 \\ \cline{2-12}
      & json & 51.64 & 0.41 & 51.39 & 51.74 & 56.33 & 7.66 & 3.82 & 7.41 & 7.39 & 45.00 \\ \hline
      
\multirow{2}{*}{GPT3.5} & text &  68.92 &	17.71 &	68.20 &	68.89 &	79.00 &  31.00 & 24.51 & 30.20 & 30.83 & 40.33 \\ \cline{2-12}
      & json & 68.44 &	17.26 &	67.72 &	67.92 &	79.00 & 30.33 &	23.92 &	29.57 &	29.29 &	40.33 \\ \hline

\multirow{2}{*}{\stackbox[c]{\textsc{orkg}-FLAN-T5$_{R0}$}} & text  & 78.64 &	28.75 &	78.33 &	78.65 & 86.33 & 71.34 & 27.75 & 70.96	& 71.41 & 81.00 \\ \cline{2-12}
& json & 80.77 & 28.03 & 80.43 & 80.53 & 88.67 & 30.93	& 27.04 &	30.55 &	30.41 &  44.67               \\ \hline                 
    \end{tabular}
  \end{adjustbox}
  \caption{Zero-shot results for T5, FLAN-T5 and GPT3.5 tested out-of-the-box to generate structured summaries versus our \textsc{orkg}-FLAN-T5$_{R0}$ model. Two answer formats plus highest \& lowest scores are contrasted. The general accuracy shows models' ability to distinguish between \textit{answerable} vs. \textit{unanswerable} contexts (details in \autoref{corpus}).
  }
  \label{zero-shot}
\end{table*}

\begin{table*}[!htb]
\centering
\begin{adjustbox}{width=1\textwidth}
	\begin{tabular}{|l|l|rrrrrrr|rrrrrrr|l|} \hline
		&         & \multicolumn{7}{|c|}{Own Test Instructions} & \multicolumn{7}{c}{Best Test Instructions} &     \\ \toprule
		&  & \stackbox[c]{\bf Disease-\\Name}  & \bf Location & \bf Date & \stackbox[c]{\bf R0-\\Value} &  \stackbox[c]{\bf \%CI-\\Values}  & \bf Method & \bf Overall & \stackbox[c]{\bf Disease-\\Name} & \bf Location & \bf Date & \stackbox[c]{\bf R0-\\Value} & \stackbox[c]{\bf \%CI-\\Values} & \bf Method & \bf Overall & \\ \hline
  
		\multirow{2}{*}{s7} & \textcolor{orange}{Exact} & 54.26 &	56.23 &	29.67 &	52.90 &	32.76 &	34.42 &	43.59 & 56.76 &	\textcolor{orange}{55.81} &	30.94 &	53.38 &	33.33 &	\textcolor{orange}{37.17} &	44.80  &   \multirow{2}{*}{s8} \\ \cline{2-16}
		& \textcolor{blue}{Partial} & 54.26 &	\textcolor{blue}{59.13} & 46.15 &	57.92 &	62.07 &	44.51 &	54.46 & 56.76 &	58.72 &	47.51 &	58.80 &	63.16 &	\textcolor{blue}{47.79} & \textcolor{blue}{55.89} &  \\ \hline
		\multirow{2}{*}{s6} & \textcolor{orange}{Exact} & 54.50 &	52.25 &	33.18 &	52.50 &	36.84 &	33.14 &	43.75 & \textcolor{orange}{58.51} &	53.11 &	35.41 &	53.00 &	\textcolor{orange}{37.84} & 33.33 &	\textcolor{orange}{45.21} & \multirow{2}{*}{s1} \\ \cline{2-16}
		& \textcolor{blue}{Partial} & 56.08 &	55.06 &	48.34 &	60.30 &	63.16 &	40.70 &	54.06 & \textcolor{blue}{60.11} &	55.93 &	49.76 &	\textcolor{blue}{61.44} & \textcolor{blue}{64.86} & 41.52 &	55.71 & \\ \hline
		\multirow{2}{*}{d3} & \textcolor{orange}{Exact} & 57.66 &	55.71 &	35.56 &	53.99 &	18.80 &	32.29 &	42.34 & 58.29 &	55.17 &	\textcolor{orange}{35.62} &	\textcolor{orange}{56.07} &	22.22 &	32.75 &	43.37 & \multirow{2}{*}{s6}  \\ \cline{2-16}
		& \textcolor{blue}{Partial} & 59.22 &	57.38 &	52.44 &	58.60 &	56.41 &	41.93 & 54.44 & 59.89 & 57.47 & \textcolor{blue}{52.97} &	61.21 &	58.12 &	42.11 & 55.42 & \\ \hline
	\end{tabular}
\end{adjustbox}
\caption{Our top three \textsc{orkg}-FLAN-T5$_{R0}$ single-task instruction-finetuned models, based on the single-instruction tuning setting in descending order of overall partial F1 for the text answer format. 1st column: models trained on SQuAD\_v2 instr. 7 (s7), SQuAD\_v2 instr. 6 (s6), and DROP instr. 3 (d3). Last column: best inference instructions.}
\label{finetuned-text}
\end{table*}

\begin{table*}[!htb]
\centering
\begin{adjustbox}{width=1\textwidth}
	\begin{tabular}{|l|l|rrrrrrr|rrrrrrr|l|} \hline
		&         & \multicolumn{7}{|c|}{Own Test Instructions} & \multicolumn{7}{c}{Best Test Instructions} &     \\ \toprule
		&  & \stackbox[c]{\bf Disease-\\Name}  & \bf Location & \bf Date & \stackbox[c]{\bf R0-\\Value} &  \stackbox[c]{\bf \%CI-\\Values}  & \bf Method & \bf Overall & \stackbox[c]{\bf Disease-\\Name} & \bf Location & \bf Date & \stackbox[c]{\bf R0-\\Value} & \stackbox[c]{\bf \%CI-\\Values} & \bf Method & \bf Overall & \\ \hline
  
		\multirow{2}{*}{d3} & \textcolor{orange}{Exact} & 55.64 &	53.04 &	32.84 &	47.62 &	24.56 &	32.64 &	41.11 & \textcolor{orange}{59.26} &	53.33 &	\textcolor{orange}{35.18} &	49.20 &	25.00 &	\textcolor{orange}{35.12} & \textcolor{orange}{42.91} & \multirow{2}{*}{s6} \\ \cline{2-16}
		& \textcolor{blue}{Partial} & 58.27 &	56.35 &	51.74 &	54.19 &	56.14 &	45.10 &	53.84 & \textcolor{blue}{61.38} &	56.67 &	\textcolor{blue}{54.27} &	\textcolor{blue}{56.95} &	55.36 &	 \textcolor{blue}{45.83} &	 \textcolor{blue}{55.28} & \\ \hline
		\multirow{2}{*}{s8} & \textcolor{orange}{Exact} & 54.08 &	53.51 &	34.91 &	48.92 &	24.56 &	30.42 &	41.10 & 56.85 &	\textcolor{orange}{54.25} &	31.88 &	\textcolor{orange}{49.53} &	\textcolor{orange}{27.27} &	31.34 &	41.89 & \multirow{2}{*}{s1} \\ \cline{2-16}
		& \textcolor{blue}{Partial} & 56.63 &	56.22 &	50.94 &	55.83 &	52.63 &	41.13 &	52.34 & 59.43 &	\textcolor{blue}{56.99} &	49.28 &	55.53 &	56.36 &	42.17 &	53.39 & \\ \hline
		\multirow{2}{*}{s10} & \textcolor{orange}{Exact} & 52.92 &	52.20 &	34.74 &	47.52 &	16.82 &	32.82 &	39.56 & 57.14 &	52.23 &	33.33 &	48.32 &	17.65 &	32.70 &	40.31 & \multirow{2}{*}{s1}  \\ \cline{2-16}
		& \textcolor{blue}{Partial} & 54.04 & 54.55 & 50.53 &	53.59 &	56.07 &	41.49 & 51.82 & 58.26 & 54.60 &	49.46 &	54.67 &	\textcolor{blue}{58.82} & 40.88 & 52.91 & \\ \hline
	\end{tabular}
\end{adjustbox}
\caption{Our top three \textsc{orkg}-FLAN-T5$_{R0}$ single-task instruction-finetuned models, based on the single-instruction tuning setting in descending order of overall partial F1 for the JSON answer. 1st column: models trained on DROP instr. 3 (d3), SQuAD\_v2 instr. 8 (s8), and SQuAD\_v2 instr. 10 (s10). Last column: best inference instructions.}
\label{finetuned-json}
\end{table*}

\section{Evaluations}
\label{eval}

\noindent{\textbf{Dataset.}} For evaluations, we created a 70\%/10\%/ 20\% split as train/dev/test sets, respectively, of the 1500 instances. The dataset comprised 1,082 train (464 ans, 618 unans), 120 dev (53 ans, 67 unans), and 300 test (135 ans, 165 unans) instances.


\noindent{\textbf{Experimental setup.}} We used a total of 18 instructions for training, with 9 instructions each from SQuAD\_v2 and DROP suitable for our task. Among these, 2 DROP instructions were formulated to prompt the LLM to generate a question from a given context. Although indirect to our task, we included them as they were relevant to obtaining capable models, but were excluded from testing. Thus for testing, we had 16 instructions (9 SQuAD and 7 DROP). For training, we had three main experimental settings based on the 18 training instructions. In the first setting, we trained 32 models (16 for text-format and 16 for JSON-format) by tuning FLAN-T5 with a single instruction for our task. Note here models were not trained for the indirect instruction. This setting tested the hypothesis that FLAN-T5 only needed one instruction to perform our task effectively since it already came instruction tuned. In the second setting, we trained two models: one using all 18 instructions with the full training data, and the other using a 50\% random sub-sample to prevent overfitting. This resulted in four models for each answer format. The third setting followed a similar approach, training two models with best SQuAD and DROP instructions based on single instruction inference results. Overall, we trained 40 models. Model hyperparamter details are in \autoref{app:hyp}. In terms of compute, all experiments were run on an NVIDIA 3090 GPU. Training took 12-15 hours on smaller datasets and 30 hours on larger datasets, while inference lasted 15-30 minutes for 300 test instances.

\noindent{\textbf{Metrics.}} We experimented in two main settings: zero-shot evaluations and single-task finetuned model evaluations. For the latter, we used recall, precision, and F1 metrics in exact and partial match settings for each of the six \textsc{orkg-R0} extraction targets and overall. In the zero-shot evaluations, where models were not guaranteed to respond with the desired structure, we treated the task as structured summarization. To evaluate these summaries, we used standard summarization ROUGE metrics~\cite{rouge} (details in \autoref{app:rouge}) instead of F1 metrics, which would require complex post-processing and could lead to misinterpretation of the model's response.

\subsection{Results and Discussion}


\noindent{\textbf{Zero-shot evaluations.}} \autoref{zero-shot} results show model's capacity in generating structured summaries per \textsc{orkg-R0}. Notably, our single-task instruction-finetuned \textsc{orkg}-FLAN-T5$_{R0}$ model surpasses its incremental predecessors T5 and FLAN-T5 with the same parameter size of 780M, as well as GPT3.5 (with 1000x more parameters at 175B), confirming the effectiveness of instruction-tuned models for single-task finetuning. Additionally, the general accuracy of the model, which distinguishes between answerable and unanswerable contexts, is significantly improved, at nearly 89\%.


\noindent{\textbf{Single-task finetuning of instruction-tuned LLM.}} From the 40 trained models, the best results were achieved in the single-instruction tuning setting, as shown in \autoref{finetuned-text} and \ref{finetuned-json} for text and JSON answers respectively. The best partial overall F1 scores were 55.89\% for text answer and 55.28\% for JSON answer. Among the 6 properties, extracting $R0$ and \%CI values was relatively easier with higher scores for text than JSON. Extracting the method and date proved to be the most challenging. Since our work builds upon the instruction-tuned FLAN-T5 model, it already possesses the capability to handle the instructions we use. Thus, the best inference instruction was not necessarily the same as the one the model was trained on. More results from the all-instruction and best-instruction models can be found in \autoref{app:res}.

\begin{figure}[!tb]
\includegraphics[width=\linewidth]{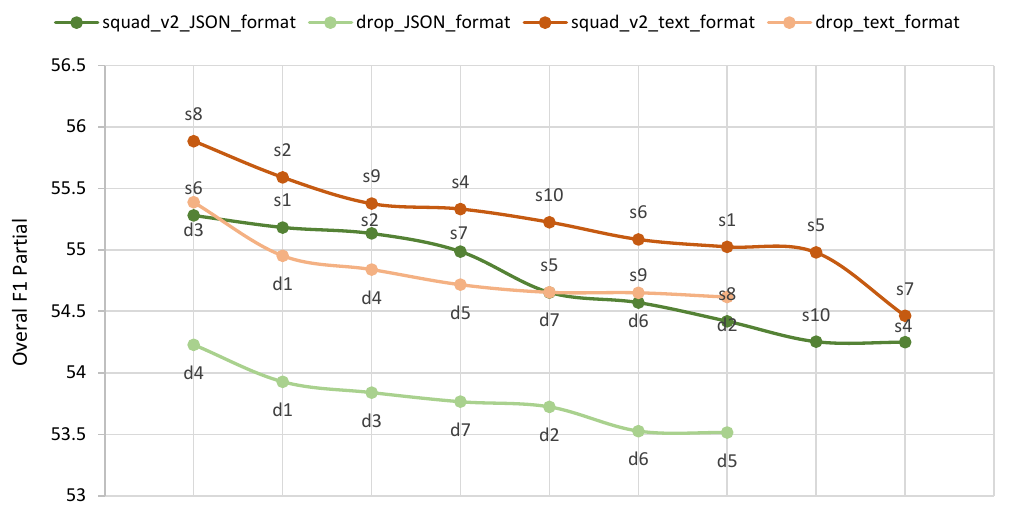}
\caption{Performances range on inference instructions.}
\label{templates-comparison}
\end{figure}

\noindent{\textbf{Impact of diverse inference instructions.}} \autoref{templates-comparison} offers a look into the inference performance differences from the best \textsc{orkg}-FLAN-T5$_{R0}$ model. As such the model shows better responses to the SQuAD (orange lines) versus DROP (green lines) inference templates in both text (darker lines) and JSON (lighter lines) answers.




\section{Error Analysis}


Based on an analysis of all the erroneous responses on the test set from our best model, we identified five main error types. They were further categorized on their impact on recall or precision. For each, mismatching (prediction, annotated label), we assigned an error type(s) and on which properties that error had an effect. The five error types are: \textit{\underline{Type 1}} is where the model answers unanswerable questions (Type 1.1) or fails to provide answers for answerable questions (Type 1.2). \textit{\underline{Type 2}} is where the model predicts values for a property and the label had no value (Type 2.1) or does not predict a value when the label had a value (Type 2.2). \textit{\underline{Type 3}} is where the model predicts either more (Type 3.1) or fewer (Type 3.2) contributions than indicated in the label. \textit{\underline{Type 4}} were inconsistencies between predicted and label values. This may include minor typographical errors (Type 4.1), not fully addressing the label values but still providing a related value in prediction (Type 4.2), including extra related information in prediction (Type 4.3), or generating totally unrelated predicted values (Type 4.4). \textit{\underline{Type 5.1}} is an invalid predicted JSON.


\paragraph{Text Response Format.}

\begin{figure}[!tb]
\includegraphics[width=\linewidth]{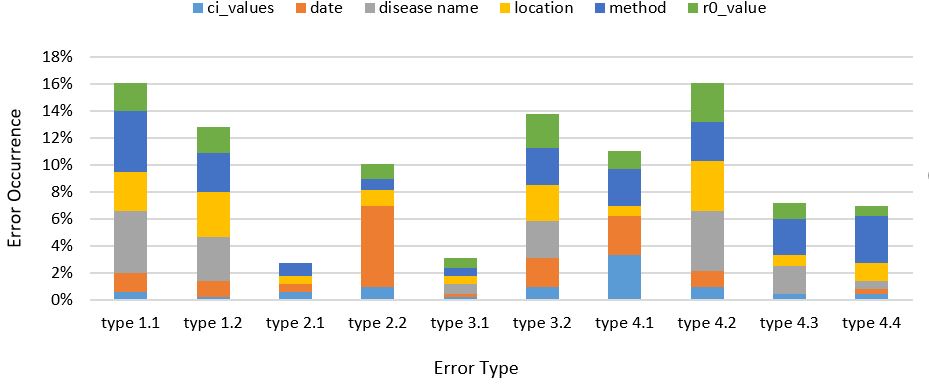}
\caption{Our best model error types for text format.}
\label{text-based-error}
\end{figure}

As shown in \autoref{text-based-error}, the most frequent errors in the text-based settings are unanswerable labels (Type 1.1) and incomplete predictions (Type 4.2). These two errors have similar distributions across properties and "method" is the most affected property overall. Type 2.2 errors significantly impact the accuracy of extracting "date" values. In contrast, Type 2.1 and Type 3.1 errors are rare, indicating the model's ability to generate property values and contributions appropriately. Typographical errors (Type 4.1) are common, particularly for "\%CI values" and "date," suggesting that normalizing label values and using a standard can improve performance in this regard.


\paragraph{JSON Response Format.}

\begin{figure}[!tb]
\includegraphics[width=\linewidth]{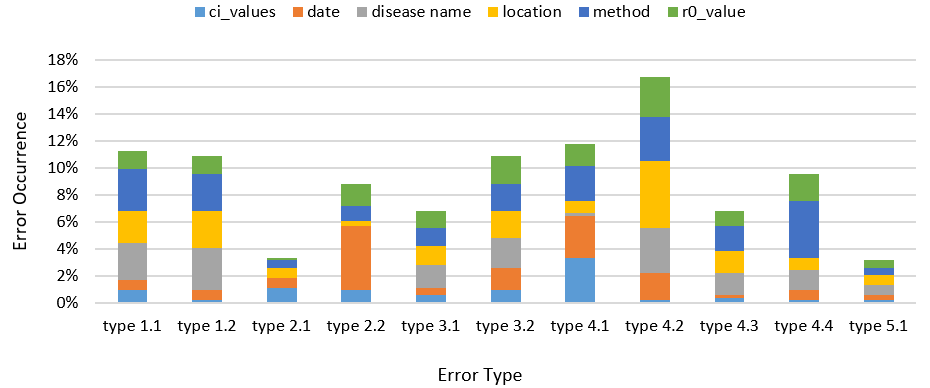}
\caption{Our best model error types for JSON format.}
\label{json-based-error}
\end{figure}

\autoref{json-based-error} shows error Type 4.2 is the primary error affecting properties, similar to text-format errors. The "method" property is the most affected overall, while "date" is particularly impacted by error Type 2.2, highlighting a common issue in JSON-based models. However, JSON models exhibit fewer errors of Type 1 (unanswerable) and instead tend to make more errors in predicting extra text (Type 2.1 and Type 3.1).


\section{Conclusions and Future Directions}
Searching scientific articles for the \href{https://scholar.google.com/scholar?hl=en&as_sdt=0%2C5&q=COVID-19+R0&btnG=}{Covid-19 R0 estimate} yields around 44,000 results. To navigate through this vast amount of information and stay up-to-date with the latest R0 estimates, is inundating for researchers. Next-generation digital libraries like ORKG are transforming this traditional paradigm by capturing machine-actionable data, enabling advanced computational tools such as \href{https://orkg.org/comparison/R44930/}{research comparisons}. LLM-powered complex IE technologies can play a crucial role in scaling scientific information extraction. We present a concrete use-case in virology, showcasing the acquisition of LLM-powered structured knowledge with the \textsc{orkg-R0} model. Our \href{https://doi.org/10.5281/zenodo.8068442}{dataset} (CC BY 4.0 license), \href{https://anonymous.4open.science/r/R0_Structured_Information_Extraction-5920/src/data/create_templated_datasets/build_templated_datasets.py}{instructions}, and \href{https://anonymous.4open.science/r/R0_Structured_Information_Extraction-5920/README.md}{source code} (MIT license) are released to support reproducibility and future research.

Our work can be seen as a flavor of meta-learning that was seminally proposed by \citet{metaicl} as the meta in-context learning paradigm. We explore meta-learning through instruction-finetuning of an instruction-tuned model, and differ in that we use a zero-shot rather than a few-shot training and testing scenario. We relegate few-shot in-context model learning to future work. While this work comprehensively evaluates the T5 class of LLMs, there are other promising LLMs like PaLM~\cite{palm}, Chinchilla~\cite{chinchilla}, and ChatGPT~\cite{gpt3,instructgpt} that can be further investigated for NLP tasks with instructions. Exploring alternative model families is a fruitful direction for future research. Additionally, model distillation~\cite{distil1,distil2,distil3,distil4} holds potential for transferring knowledge from large teacher models to smaller, efficient student models. This approach holds promise, particularly in scenarios where single-task tuned models are desired, as we propose in this study. 

\section*{Limitations}

This section  presents a discussion of the limitations w.r.t. the two main facets of this work: structured scholarly knowledge publishing (paragraph I) and LLM scaling experiments for single-task instruction finetuning (paragraph II).  

\paragraph{I. Structured Scholarly Knowledge Publishing}

This work proposes the \textsc{orkg-R0} model that records a fine-grained structured representation of the salient facets of a research contribution on the specific research problem of investigating the R0 number of infectious diseases. For such popular research use-cases in the community, e.g., capturing Leaderboards in the empirical AI research as Task, Dataset, Metric, and Score~\cite{kabongo2021automated,kabongo2023orkg,kabongo2023zero}, as another example apart from the one we address in this work, a current limitation that such a contribution-centric fine-grained structured scholarly knowledge publishing model faces is it's \textit{adoption and standardization}. The widespread adoption of the semantic scholarly knowledge publishing model is still in its early stages, and achieving consensus on standard formats, ontologies, and metadata remains a challenge. This lack of standardization can hinder interoperability and limit the accessibility of knowledge across different platforms and communities. To overcome this limitation, i.e. to realize this vision of the publishing of fine-grained structured scholarly contributions to better assist researchers to stay on track with research progress many more collaborative advocacy and community-building efforts would need to be set in place. The trajectory, however, looks promising. The ORKG since its inception in 2018 currently has a knowledge base of roughly 41k structured contributions. More stats here \url{https://orkg.org/stats}. In addition, yearly paid community curation grants are run inviting researchers from various disciplines to help curate a high-quality knowledge graph (\url{https://orkg.org/about/28/Curation_Grants}). Finally, the ORKG has initiated collaborations with various conferences and journals that ask authors to submit research comparisons of their work versus related work to help expedite the peer-review process. E.g., see the last point in the Author Guidelines in the SEMANTiCS 2023 call for papers \url{https://2023-eu.semantics.cc/page/cfp_rev_rep}. To this end, the platform is integrated with content creator anonymization features to support double-blind review protocols. More information here \url{https://orkg.org/about/22/Conferences_and_Journals}.

As a second limitation of semantic publishing, the ORKG is designed to be a next-generation digital library that supports fine-grained scholarly knowledge publishing stored as a large-scale knowledge graph in the backend~\cite{jaradeh2019open}. It is also amenable to be published in the Linked Open Data (LOD) Cloud \url{https://lod-cloud.net/}. Thus it follows the best practices laid out in \citeposs{berners2001semantic} the Semantic Web. As such the engineering of this platform entails a high degree of \textit{technical complexity} compared with the traditional PDF-based publishing platforms. Implementing and maintaining the infrastructure required for semantic publishing models can be technically complex and resource-intensive. It requires expertise in semantic technologies, data management, and ontological engineering. Nevertheless, the ORKG platform supports the integration of widgets for its various features in other platforms. This would lower the technical entrance barrier for other publishers to also support the semantic publishing of scientific contributions.


\paragraph{II. Scaling Single-Task Instruction-tuning of LLMs}

This work has investigated the moderate-sized FLAN-T5 Large model with 780M parameters. Prior work reported: ``we see that increasing model scale by an order of magnitude (i.e., 8B -> 62B or 62B -> 540B) improves performance substantially for both finetuned and non-finetuned models''~\cite{flan-t5}. Borrowing insights from the earlier experiments on scaling models, potentially, a single-task finetuned model performance could be boosted if larger scale models were used. This aspect while not analyzed in this work is relegated to future work. However, a more practically viable option would not just be additional scaling investigations, but these combined with model distillation~\cite{distil1}.

\bibliography{custom}

\appendix

\section{Instructions: Qualitative Examples}
\label{app:temp}

In this section, we elicit each of the instructions that were considered in this work as formulated in the FLAN 2022 Collection for the SQuAD\_v2 and DROP datasets.

\subsection{The Stanford Question Answering Dataset (SQuAD\_v2)}

\paragraph{Instruction 1:}
\mbox{}\\

\textbf{title}: Estimating the serial interval of the novel coronavirus disease (COVID-19) based on the public surveillance data in Shenzhen, China, from 19 January to 22 February 2020

\textbf{context}: The novel coronavirus disease (COVID-19) poses a serious threat to global public health and economics. Serial interval (SI), time between the onset of symptoms of a primary case and a secondary case, is a key epidemiological parameter. We estimated SI of COVID-19 in Shenzhen, China based on 27 ...

\textbf{question}: What are the values for the following properties of the basic reproduction number estimate (R0): disease name, location, date, R0 value, \%CI values, and method?

\textbf{Instruction}:

{\{title\}}:\textbackslash n\textbackslash n\{context\}\textbackslash n\textbackslash n Please answer a question about this article. If the question is unanswerable, say "unanswerable". \{question\}

\paragraph{Instruction 2:}
\mbox{}\\

\textbf{context}: 
Estimating the serial interval of the novel coronavirus disease (COVID-19) based on the public surveillance data in Shenzhen, China, from 19 January to 22 February 2020

The novel coronavirus disease (COVID-19) poses a serious threat to global public health and economics. Serial interval (SI), time between the onset of symptoms of a primary case and a secondary case, is a key epidemiological parameter. We estimated SI of COVID-19 in Shenzhen, China based on 27 ...

\textbf{question}: What are the values for the following properties of the basic reproduction number estimate (R0): disease name, location, date, R0 value, \%CI values, and method?

\textbf{Instruction}:
Read this and answer the question. If the question is unanswerable, say "unanswerable".\textbackslash n\textbackslash n\{context\}\textbackslash n\textbackslash n\{question\}

\paragraph{Instruction 3:}
\mbox{}\\

\noindent{\textit{This instruction is omitted in this work.}}

\textbf{Instruction}:
(What is a question about this article? If the question is unanswerable, say "unanswerable"),\textbackslash n\{context\}\textbackslash n\{question\} 

\paragraph{Instruction 4:}
\mbox{}\\

\textbf{context}: 
Estimating the serial interval of the novel coronavirus disease (COVID-19) based on the public surveillance data in Shenzhen, China, from 19 January to 22 February 2020

The novel coronavirus disease (COVID-19) poses a serious threat to global public health and economics. Serial interval (SI), time between the onset of symptoms of a primary case and a secondary case, is a key epidemiological parameter. We estimated SI of COVID-19 in Shenzhen, China based on 27 ...

\textbf{question}: What are the values for the following properties of the basic reproduction number estimate (R0): disease name, location, date, R0 value, \%CI values, and method?

\textbf{Instruction}:
\{context\}\textbackslash n\{question\} (If the question is unanswerable, say "unanswerable")

\paragraph{Instruction 5:}
\mbox{}\\

\textbf{context}: 
Estimating the serial interval of the novel coronavirus disease (COVID-19) based on the public surveillance data in Shenzhen, China, from 19 January to 22 February 2020

The novel coronavirus disease (COVID-19) poses a serious threat to global public health and economics. Serial interval (SI), time between the onset of symptoms of a primary case and a secondary case, is a key epidemiological parameter. We estimated SI of COVID-19 in Shenzhen, China based on 27...

\textbf{question}: What are the values for the following properties of the basic reproduction number estimate (R0): disease name, location, date, R0 value, \%CI values, and method?

\textbf{Instruction}:
\{context\}\textbackslash n Try to answer this question if possible (otherwise reply "unanswerable"):\{question\}

\paragraph{Instruction 6:}
\mbox{}\\

\textbf{context}: 
Estimating the serial interval of the novel coronavirus disease (COVID-19) based on the public surveillance data in Shenzhen, China, from 19 January to 22 February 2020

The novel coronavirus disease (COVID-19) poses a serious threat to global public health and economics. Serial interval (SI), time between the onset of symptoms of a primary case and a secondary case, is a key epidemiological parameter. We estimated SI of COVID-19 in Shenzhen, China based on 27...

\textbf{question}: What are the values for the following properties of the basic reproduction number estimate (R0): disease name, location, date, R0 value, \%CI values, and method?

\textbf{Instruction}:
\{context\}\textbackslash n If it is possible to answer this question, answer it for me (else, reply "unanswerable"): \{question\}

\paragraph{Instruction 7:}
\mbox{}\\

\textbf{context}: 
Estimating the serial interval of the novel coronavirus disease (COVID-19) based on the public surveillance data in Shenzhen, China, from 19 January to 22 February 2020

The novel coronavirus disease (COVID-19) poses a serious threat to global public health and economics. Serial interval (SI), time between the onset of symptoms of a primary case and a secondary case, is a key epidemiological parameter. We estimated SI of COVID-19 in Shenzhen, China based on 27...

\textbf{question}: What are the values for the following properties of the basic reproduction number estimate (R0): disease name, location, date, R0 value, \%CI values, and method?

\textbf{Instruction}:
\{context\}\textbackslash n \textbackslash n Answer this question, if possible (if impossible, reply "unanswerable"): \{question\}

\paragraph{Instruction 8:}
\mbox{}\\

\textbf{context}: 
Estimating the serial interval of the novel coronavirus disease (COVID-19) based on the public surveillance data in Shenzhen, China, from 19 January to 22 February 2020

The novel coronavirus disease (COVID-19) poses a serious threat to global public health and economics. Serial interval (SI), time between the onset of symptoms of a primary case and a secondary case, is a key epidemiological parameter. We estimated SI of COVID-19 in Shenzhen, China based on 27...

\textbf{question}: What are the values for the following properties of the basic reproduction number estimate (R0): disease name, location, date, R0 value, \%CI values, and method?

\textbf{Instruction}:
Read this: \{context\}\textbackslash n \textbackslash n \{question\} \textbackslash n What is the answer? (If it cannot be answered, return "unanswerable")

\paragraph{Instruction 9:}
\mbox{}\\

\textbf{context}: 
Estimating the serial interval of the novel coronavirus disease (COVID-19) based on the public surveillance data in Shenzhen, China, from 19 January to 22 February 2020

The novel coronavirus disease (COVID-19) poses a serious threat to global public health and economics. Serial interval (SI), time between the onset of symptoms of a primary case and a secondary case, is a key epidemiological parameter. We estimated SI of COVID-19 in Shenzhen, China based on 27...

\textbf{question}: What are the values for the following properties of the basic reproduction number estimate (R0): disease name, location, date, R0 value, \%CI values, and method?

\textbf{Instruction}:
Read this: \{context\}\textbackslash n Now answer this question, if there is an answer (If it cannot be answered, return "unanswerable"): \{question\}

\paragraph{Instruction 10:}
\mbox{}\\

\textbf{context}: 
Estimating the serial interval of the novel coronavirus disease (COVID-19) based on the public surveillance data in Shenzhen, China, from 19 January to 22 February 2020

The novel coronavirus disease (COVID-19) poses a serious threat to global public health and economics. Serial interval (SI), time between the onset of symptoms of a primary case and a secondary case, is a key epidemiological parameter. We estimated SI of COVID-19 in Shenzhen, China based on 27...

\textbf{question}: What are the values for the following properties of the basic reproduction number estimate (R0): disease name, location, date, R0 value, \%CI values, and method?

\textbf{Instruction}:
\{context\}\textbackslash n Is there an answer to this question (If it cannot be answered, say "unanswerable"): \{question\}

\subsection{Discrete Reasoning over Paragraphs (DROP) Dataset}

\paragraph{Instruction 1:}
\mbox{}\\

\textbf{context}: 
Estimating the serial interval of the novel coronavirus disease (COVID-19) based on the public surveillance data in Shenzhen, China, from 19 January to 22 February 2020

The novel coronavirus disease (COVID-19) poses a serious threat to global public health and economics. Serial interval (SI), time between the onset of symptoms of a primary case and a secondary case, is a key epidemiological parameter. We estimated SI of COVID-19 in Shenzhen, China based on 27...

\textbf{question}: What are the values for the following properties of the basic reproduction number estimate (R0): disease name, location, date, R0 value, \%CI values, and method?

\textbf{Instruction}:
Answer based on context: \textbackslash n \textbackslash n\{context\}\textbackslash n \textbackslash n \{question\}

\paragraph{Instruction 2:}
\mbox{}\\

\textbf{context}: 
Estimating the serial interval of the novel coronavirus disease (COVID-19) based on the public surveillance data in Shenzhen, China, from 19 January to 22 February 2020

The novel coronavirus disease (COVID-19) poses a serious threat to global public health and economics. Serial interval (SI), time between the onset of symptoms of a primary case and a secondary case, is a key epidemiological parameter. We estimated SI of COVID-19 in Shenzhen, China based on 27...

\textbf{question}: What are the values for the following properties of the basic reproduction number estimate (R0): disease name, location, date, R0 value, \%CI values, and method?

\textbf{Instruction}:
\{context\}\textbackslash n \textbackslash n Answer this question based on the article: \{question\}

\paragraph{Instruction 3:}
\mbox{}\\

\textbf{context}: 
Estimating the serial interval of the novel coronavirus disease (COVID-19) based on the public surveillance data in Shenzhen, China, from 19 January to 22 February 2020

The novel coronavirus disease (COVID-19) poses a serious threat to global public health and economics. Serial interval (SI), time between the onset of symptoms of a primary case and a secondary case, is a key epidemiological parameter. We estimated SI of COVID-19 in Shenzhen, China based on 27...

\textbf{question}: What are the values for the following properties of the basic reproduction number estimate (R0): disease name, location, date, R0 value, \%CI values, and method?

\textbf{Instruction}:
\{context\}\textbackslash n \textbackslash n \{question\}

\paragraph{Instruction 4:}
\mbox{}\\

\textbf{context}: 
Estimating the serial interval of the novel coronavirus disease (COVID-19) based on the public surveillance data in Shenzhen, China, from 19 January to 22 February 2020

The novel coronavirus disease (COVID-19) poses a serious threat to global public health and economics. Serial interval (SI), time between the onset of symptoms of a primary case and a secondary case, is a key epidemiological parameter. We estimated SI of COVID-19 in Shenzhen, China based on 27...

\textbf{question}: What are the values for the following properties of the basic reproduction number estimate (R0): disease name, location, date, R0 value, \%CI values, and method?

\textbf{Instruction}:
\{context\}\textbackslash n Answer this question: \{question\}

\paragraph{Instruction 5:}
\mbox{}\\

\textbf{context}: 
Estimating the serial interval of the novel coronavirus disease (COVID-19) based on the public surveillance data in Shenzhen, China, from 19 January to 22 February 2020

The novel coronavirus disease (COVID-19) poses a serious threat to global public health and economics. Serial interval (SI), time between the onset of symptoms of a primary case and a secondary case, is a key epidemiological parameter. We estimated SI of COVID-19 in Shenzhen, China based on 27...

\textbf{question}: What are the values for the following properties of the basic reproduction number estimate (R0): disease name, location, date, R0 value, \%CI values, and method?

\textbf{Instruction}:
Read this article and answer this question \{context\}\textbackslash n \{question\}

\paragraph{Instruction 6:}
\mbox{}\\

\textbf{context}: 
Estimating the serial interval of the novel coronavirus disease (COVID-19) based on the public surveillance data in Shenzhen, China, from 19 January to 22 February 2020

The novel coronavirus disease (COVID-19) poses a serious threat to global public health and economics. Serial interval (SI), time between the onset of symptoms of a primary case and a secondary case, is a key epidemiological parameter. We estimated SI of COVID-19 in Shenzhen, China based on 27...

\textbf{question}: What are the values for the following properties of the basic reproduction number estimate (R0): disease name, location, date, R0 value, \%CI values, and method?

\textbf{Instruction}:
\{context\}\textbackslash n \textbackslash n Based on the above article, answer a question. \{question\}

\paragraph{Instruction 7:}
\mbox{}\\

\textbf{context}: 
Estimating the serial interval of the novel coronavirus disease (COVID-19) based on the public surveillance data in Shenzhen, China, from 19 January to 22 February 2020

The novel coronavirus disease (COVID-19) poses a serious threat to global public health and economics. Serial interval (SI), time between the onset of symptoms of a primary case and a secondary case, is a key epidemiological parameter. We estimated SI of COVID-19 in Shenzhen, China based on 27...

\textbf{question}: What are the values for the following properties of the basic reproduction number estimate (R0): disease name, location, date, R0 value, \%CI values, and method?

\textbf{Instruction}:
Context: \{context\}\textbackslash n \textbackslash n Question: \{question\}\textbackslash n \textbackslash n Answer:

\paragraph{Instruction 8:}
\mbox{}\\

\noindent{\textit{This instruction is omitted in this work.}}

\textbf{Instruction}:
Write an article that answers the following question: \{question\}

\paragraph{Instruction 9:}
\mbox{}\\

\noindent{\textit{Note single-instruction finetuned models were not trained on this instruction. This instruction was only used in the all-instruction training setting.}}

\textbf{context}: 
Estimating the serial interval of the novel coronavirus disease (COVID-19) based on the public surveillance data in Shenzhen, China, from 19 January to 22 February 2020

The novel coronavirus disease (COVID-19) poses a serious threat to global public health and economics. Serial interval (SI), time between the onset of symptoms of a primary case and a secondary case, is a key epidemiological parameter. We estimated SI of COVID-19 in Shenzhen, China based on 27...

\textbf{question}: What are the values for the following properties of the basic reproduction number estimate (R0): disease name, location, date, R0 value, \%CI values, and method?

\textbf{Instruction}:
Write a question about the following article: \{context\}

\paragraph{Instruction 10:}
\mbox{}\\

\noindent{\textit{Note single-instruction finetuned models were not trained on this instruction. This instruction was only used in the all-instruction training setting.}}

\textbf{context}: 
Estimating the serial interval of the novel coronavirus disease (COVID-19) based on the public surveillance data in Shenzhen, China, from 19 January to 22 February 2020

The novel coronavirus disease (COVID-19) poses a serious threat to global public health and economics. Serial interval (SI), time between the onset of symptoms of a primary case and a secondary case, is a key epidemiological parameter. We estimated SI of COVID-19 in Shenzhen, China based on 27...

\textbf{question}: What are the values for the following properties of the basic reproduction number estimate (R0): disease name, location, date, R0 value, \%CI values, and method?

\textbf{Instruction}:
\{context\}\textbackslash n \textbackslash n Ask a question about this article.

\section{\textsc{orkg-R0} for the FLAN Collection}

In this section, we discuss the relation of our complex IE task formulated as \textsc{orkg-R0} to the task types already in the FLAN collection~\citeyearpar{flan,flan-collection} as a new candidate for inclusion. As mentioned earlier, FLAN has 12 task type clusters of 63 datasets. Two of which are reading comprehension (RC) and struct-to-text, among others. In this respect, our task could either be considered part of the RC task or as a new task type i.e. text-to-struct. In an RC task, e.g. SQuAD~\citeyearpar{squad-v1}, a context passage is provided along with a question to test comprehension. Our complex IE task is similar, where given a scholarly paper's title and abstract as context, the machine must generate a structured summary by understanding the context and assigning applicable extracted values for the \textsc{orkg-R0} properties. Furthermore, the model must also create \textsc{orkg-R0} clusters for abstracts reporting multiple contributions.\footnote{Note, there is a subtle difference between RC and the related question-answering (QA) task type. In QA, complex IE would require breaking down the RC extraction target into multiple questions, such as the disease name or the reported location, etc., unlike in RC.} 
Otherwise, it could be introduced into the FLAN collection as a new task type called text-to-struct. As such, for instance, the WebNLG~\cite{webnlg} or DART~\cite{dart} datasets in the struct-to-text cluster, seek to convert structured data in \href{https://www.w3.org/TR/rdf11-concepts/}{RDF} to text. Notably, our task is its direct inverse which seeks to obtain structured property-value tuples which can easily be represented in RDF syntax. 

\section{Our Experimental Hyperparamters}
\label{app:hyp}
We had different training experimental settings to train on different datasets with different sizes (single-instruction model tuning, all-instructions model tuning, all-instructions model tuning with 50\% subsampled training data, best-instructions model tuning, and best-instructions model tuning with 50\% subsampled training data).

The hyperparameters are: batch size and number of training epochs, which differ based on each dataset group mentioned above. the batch size was either 32 or 16 and the number of epochs were one of 10, 15, 20, and 30 values. In all settings we used early stopping which stops the training if the "Overall Partial F1" score dose not improve at least 0.1\% after completing 10 consecutive training epochs. For all settings we used AdafactorSchedule and Adafactor optimizer~\cite{shazeer2018adafactor} with scale\_parameter=True, relative\_step=True, warmup\_init=True, lr=None, which is one of the combinations working well according to the community for T5 finetuning.

The evaluations were done on each epoch on the dev set and we kept two best (the one maximizing the "Overall Partial F1" score) and last checkpoints in each model training process to then use for inference on test set.

\begin{table*}[!htb]
\centering
\begin{adjustbox}{width=1\textwidth}
\begin{tabular}{lllllllllllllllllll}
\hline
\multicolumn{1}{c}{\multirow{2}{*}{}} &
  \multicolumn{1}{c}{\multirow{2}{*}{}} &
  \multicolumn{1}{c}{\multirow{2}{*}{}} &
  \multicolumn{7}{c}{All Data} &
  \multicolumn{1}{c}{\multirow{2}{*}{}} &
  \multicolumn{1}{c}{\multirow{2}{*}{}} &
  \multicolumn{7}{c}{Data From Random Selection of Templates} \\ \cline{4-10} \cline{13-19} \\
  \multicolumn{1}{c}{} &
  \multicolumn{1}{c}{Template} &
  \multicolumn{1}{c}{Match Type} &
  \stackbox[c]{Disease\\-Name} &
   Location &
   Date &
  \stackbox[c]{R0\\-Value} &
  \stackbox[c]{CI \\-\% Values} &
   Method &
   Overall &
  \multicolumn{1}{c}{Template} &
  \multicolumn{1}{c}{Match Type} &
  \stackbox[c]{Disease\\-Name} &
   Location &
   Date &
  \stackbox[c]{R0\\-Value} &
  \stackbox[c]{CI \\-\% Values} &
   Method &
   Overall \\
  \hline
\multirow{4}{*}{Top 2 Highest}  & \multirow{2}{*}{s1} & Exact   &  54.24 &	52.12 &	21.51 &	47.84 &	13.59 &	33.96 &	37.26 &
 \multirow{2}{*}{d7} & Exact   &  54.88 &	51.69	& 33.48 &	49.84 &	33.06 &	33.43 &	42.76
 \\ \cline{3-10} \cline{12-19} 
                             &                             & Partial &  54.80 &	53.94 &	38.71 &	55.22 &	54.37 &	44.65 &	50.35 &                            & Partial &  55.41 &	54.49 &	48.46 &	56.26 &	57.85 &	40.47 &	52.38  \\ \cline{2-19} 
                             & \multirow{2}{*}{d6} & Exact   & 53.52 &	51.81 &	21.51 &	47.84 &	13.59 &	33.23 &	36.96 & \multirow{2}{*}{d1} & Exact   &  54.69 &	51.70 &	29.60 &	50.16 &	36.67 &	32.14 &	42.53  \\ \cline{3-10} \cline{12-19} 
                             &                             & Partial &  54.08 &	53.61 &	37.63 &	55.29 &	54.37 &	43.89 &	49.89 &                            & Partial &  55.23 &	55.11 &	43.95 &	56.50 &	58.33 &	39.29 &	51.66 \\ \hline
\multirow{4}{*}{Top 2 Lowest} & \multirow{2}{*}{d4} & Exact   &  53.22 &	51.65 &	20.32 &	46.86 &	13.46 &	33.02 &	36.47
 & \multirow{2}{*}{s6} & Exact   &  56.02 &	47.00 &	27.56 &	45.98 &	36.07 &	31.06 &	40.63 \\ \cline{3-10} \cline{12-19} 
                             &                             & Partial &  53.78 &	53.45 &	36.36 &	54.62 &	53.85 &	42.99 &	49.25 &                             & Partial &  56.51 &	50.13 &	40.94 &	51.31 &	55.74 &	38.15 &	48.93 \\ \cline{2-19} 
                             & \multirow{2}{*}{s8} & Exact   &  53.22 &	52.25 &	19.35 &	46.71 &	13.59 &	33.64 & 36.51
 & \multirow{2}{*}{d4} & Exact   &  52.58 &	47.67 &	27.23 &	47.13 &	36.07 &	32.57 &	40.56  \\ \cline{3-10} \cline{12-19} 
                             &                             & Partial &  53.78 &	53.45 &	34.41 &	54.19 &	54.37 &	44.24 &	49.15
  &                             & Partial &  53.09 &	50.96 &	41.70 &	52.19 &	55.74 &	38.29 &	48.79
  \\ \hline
\end{tabular}
\end{adjustbox}
\caption{Top two highest and lowest inference results by \textsc{orkg}-FLAN-T5$_{R0}$ all-instructions and all-instructions with 50\% subsampled finetuned models, in descending order of overall partial F1 for the text answer. template column: inference instructions. SQuAD\_v2 instr. 1 (s1), DROP instr. 6 (d6), DROP instr. 4 (d4), SQuAD\_v2 instr. 8 (s8), DROP instr. 7 (d7), DROP instr. 1 (d1),  SQuAD\_v2 instr. 6 (s6), and DROP instr. 4 (d4).}
\label{all-instruct-text}
\end{table*}

\begin{table*}[!htb]
\centering
\begin{adjustbox}{width=1\textwidth}
\begin{tabular}{lllllllllllllllllll}
\hline
\multicolumn{1}{c}{\multirow{2}{*}{}} &
  \multicolumn{1}{c}{\multirow{2}{*}{}} &
  \multicolumn{1}{c}{\multirow{2}{*}{}} &
  \multicolumn{7}{c}{All Data} &
  \multicolumn{1}{c}{\multirow{2}{*}{}} &
  \multicolumn{1}{c}{\multirow{2}{*}{}} &
  \multicolumn{7}{c}{Data From Random Selection of Templates} \\ \cline{4-10} \cline{13-19} \\
  \multicolumn{1}{c}{} &
  \multicolumn{1}{c}{Template} &
  \multicolumn{1}{c}{Match Type} &
  \stackbox[c]{Disease\\-Name} &
   Location &
   Date &
  \stackbox[c]{R0\\-Value} &
  \stackbox[c]{CI \\-\% Values} &
   Method &
   Overall &
  \multicolumn{1}{c}{Template} &
  \multicolumn{1}{c}{Match Type} &
  \stackbox[c]{Disease\\-Name} &
   Location &
   Date &
  \stackbox[c]{R0\\-Value} &
  \stackbox[c]{CI \\-\% Values} &
   Method &
   Overall \\
  \hline
\multirow{4}{*}{Top 2 Highest}  & \multirow{2}{*}{s5} & Exact   & 51.25 &	48.94 &	29.03 &	41.97 &	13.59 &	27.04 &	35.38
 &
 \multirow{2}{*}{d4} & Exact   &  56.27 & 47.76 & 31.02 & 49.33	& 22.64 &	32.91 &	40.06 \\ \cline{3-10} \cline{12-19} 
                             &                             & Partial &  53.48 &	50.15 &	44.09 &	49.89 &	54.37 &	35.85 &	48.06
 &                            & Partial & 56.82 &	50.75 &	45.99 &	56.19 &	54.72 &	42.41 &	51.27  \\ \cline{2-19} 
                             & \multirow{2}{*}{d2} & Exact   & 50.14 &	48.94 &	26.88 &	41.97 &	13.59 &	27.67 &	34.93 
 & \multirow{2}{*}{d6} & Exact   & 56.27 &	47.76 &	31.02 &	50.00 &	22.64 &	32.38 &	40.08
 \\ \cline{3-10} \cline{12-19} 
                             &                             & Partial &  52.37 &	50.15 &	44.09 &	49.68 &	54.37 &	36.48 &	47.95
 &                            & Partial &  56.82 &	50.75 &	45.99 &	56.32 &	54.72 &	41.90 &	51.20  \\ \hline
\multirow{4}{*}{Top 2 Lowest} & \multirow{2}{*}{s1} & Exact   &  50.70 &	47.13 &	25.81 &	42.11 &	13.59 &	25.79 &	34.25

 & \multirow{2}{*}{s8} & Exact   &  54.55 &	47.06 &	32.46 &	49.01 &	22.43 &	32.50 &	39.72
  \\ \cline{3-10} \cline{12-19} 
                             &                             & Partial &  52.92 &	48.34 &	44.09 &	49.02 &	54.37 &	33.96 &	47.21 &                             & Partial &  55.10 &	50.00 &	46.07 &	55.14 &	50.47 &	41.88 &	49.88
  \\ \cline{2-19} 
                             & \multirow{2}{*}{d1} & Exact   &  50.42 &	47.42 &	25.95 &	41.58 &	13.73 &	25.95 &	34.24

 & \multirow{2}{*}{s9} & Exact   &  54.14 &	47.34 &	32.46 &	49.83 &	20.75 &	31.97 &	39.47
  \\ \cline{3-10} \cline{12-19} 
                             &                             & Partial & 52.66	& 49.24	& 44.32 &	47.83 &	52.94 &	34.81 &	47.06 

  &                             & Partial & 54.70 &	50.30 &	46.07 &	55.75 &	49.06 &	41.38 &	49.64
 \\ \hline
\end{tabular}
\end{adjustbox}
\caption{Top two highest and lowest inference results by \textsc{orkg}-FLAN-T5$_{R0}$ all-instructions and all-instructions with 50\% subsampled finetuned models, in descending order of overall partial F1 for the JSON answer. template column: inference instructions. SQuAD\_v2 instr. 5 (s5), DROP instr. 2 (d2), SQuAD\_v2 instr. 1 (s1), DROP instr. 1 (d1), DROP instr. 4 (d4), DROP instr. 6 (d6), SQuAD\_v2 instr. 8 (s8),  SQuAD\_v2 instr. 9 (s9).}
\label{all-instruct-json}
\end{table*}

\begin{table*}[!htb]
\centering
\begin{adjustbox}{width=1\textwidth}
\begin{tabular}{lllllllllllllllllll}
\hline
\multicolumn{1}{c}{\multirow{2}{*}{}} &
  \multicolumn{1}{c}{\multirow{2}{*}{}} &
  \multicolumn{1}{c}{\multirow{2}{*}{}} &
  \multicolumn{7}{c}{All Data} &
  \multicolumn{1}{c}{\multirow{2}{*}{}} &
  \multicolumn{1}{c}{\multirow{2}{*}{}} &
  \multicolumn{7}{c}{Data From Random Selection of Templates} \\ \cline{4-10} \cline{13-19} \\
  \multicolumn{1}{c}{} &
  \multicolumn{1}{c}{Template} &
  \multicolumn{1}{c}{Match Type} &
  \stackbox[c]{Disease\\-Name} &
   Location &
   Date &
  \stackbox[c]{R0\\-Value} &
  \stackbox[c]{CI \\-\% Values} &
   Method &
   Overall &
  \multicolumn{1}{c}{Template} &
  \multicolumn{1}{c}{Match Type} &
  \stackbox[c]{Disease\\-Name} &
   Location &
   Date &
  \stackbox[c]{R0\\-Value} &
  \stackbox[c]{CI \\-\% Values} &
   Method &
   Overall \\
  \hline
\multirow{4}{*}{Top 2 Highest}  & \multirow{2}{*}{s2} & Exact   &  49.21 &	54.85 &	30.00 &	49.20 &	22.81 &	32.35 &	39.79  &
 \multirow{2}{*}{s6} & Exact   &   48.04 &	47.15 &	24.88 &	41.59 &	19.42 &	23.18 &	34.16	 \\ \cline{3-10} \cline{12-19} 
                             &                             & Partial &   50.26 &	57.06 &	51.00 &	54.35 &	52.63 &	44.12 &	51.73

 &                            & Partial &  48.53 &	49.86 &	38.28 & 49.12 &	54.37 &	38.27 &	46.62

  \\ \cline{2-19} 
                             & \multirow{2}{*}{d6} & Exact   &  49.10 &	53.66 &	31.84 &	49.22 &	23.42 &	31.70 &	39.89

 & \multirow{2}{*}{d3} & Exact   &  47.62 &	46.19 &	26.92 &	41.92 &	18.35 & 21.47 &	33.87

 \\ \cline{3-10} \cline{12-19} 
                             &                             & Partial &   50.65 &	55.83 &	47.76 &	54.55 &	54.05 &	43.23 &	51.15

 &                            & Partial &  48.10 &	48.82 &	41.35 &	48.28 &	55.05 &	36.13 &	46.48

  \\ \hline
\multirow{4}{*}{Top 2 Lowest} & \multirow{2}{*}{s9} & Exact   &   48.04 &	52.05 &	32.16 &	49.21 &	23.42 &	32.56 &	39.65

 & \multirow{2}{*}{s8} & Exact   &   47.39 &	46.35 &	21.72 &	41.18 &	17.24 &	21.47 &	32.60

  \\ \cline{3-10} \cline{12-19} 
                             &                             & Partial &  49.61 &	54.25 &	47.24 &	54.43 &	54.05 &	43.60 &	50.66
	 
 &                             & Partial &   47.87 &	48.96 &	34.39 &	48.57 &	51.72 &	35.08 &	44.50

  \\ \cline{2-19} 
                             & \multirow{2}{*}{s1} & Exact   &   47.92 &	51.37 &	30.00 &	47.80 &	23.01 &	32.56 &	38.84

 & \multirow{2}{*}{s9} & Exact   &   46.90 &	44.44 &	22.33 &	40.00 &	16.39 &	21.88 &	32.07

  \\ \cline{3-10} \cline{12-19} 
                             &                             & Partial &  49.48 &	53.55 &	45.00 &	53.28 &	53.10 &	43.60 &	49.80

  &                             & Partial &  47.36 &	46.97 &	34.42 &	45.99 &	47.54 & 35.11 &	43.01

 \\ \hline
\end{tabular}
\end{adjustbox}
\caption{Top two highest and lowest inference results by \textsc{orkg}-FLAN-T5$_{R0}$ best-instructions and best-instructions with 50\% subsampled finetuned models, in descending order of overall partial F1 for the text answer. template column: inference instructions. SQuAD\_v2 instr. 2 (s2), DROP instr. 6 (d6), SQuAD\_v2 instr. 9 (s9), SQuAD\_v2 instr. 1 (s1), SQuAD\_v2 instr. 6 (s6), DROP instr. 3 (d3),  SQuAD\_v2 instr. 8 (s8), and SQuAD\_v2 instr. 9 (s9).}
\label{best-instruct-text}
\end{table*}

\begin{table*}[!htb]
\centering
\begin{adjustbox}{width=1\textwidth}
\begin{tabular}{lllllllllllllllllll}
\hline
\multicolumn{1}{c}{\multirow{2}{*}{}} &
  \multicolumn{1}{c}{\multirow{2}{*}{}} &
  \multicolumn{1}{c}{\multirow{2}{*}{}} &
  \multicolumn{7}{c}{All Data} &
  \multicolumn{1}{c}{\multirow{2}{*}{}} &
  \multicolumn{1}{c}{\multirow{2}{*}{}} &
  \multicolumn{7}{c}{Data From Random Selection of Templates} \\ \cline{4-10} \cline{13-19} \\
  \multicolumn{1}{c}{} &
  \multicolumn{1}{c}{Template} &
  \multicolumn{1}{c}{Match Type} &
  \stackbox[c]{Disease\\-Name} &
   Location &
   Date &
  \stackbox[c]{R0\\-Value} &
  \stackbox[c]{CI \\-\% Values} &
   Method &
   Overall &
  \multicolumn{1}{c}{Template} &
  \multicolumn{1}{c}{Match Type} &
  \stackbox[c]{Disease\\-Name} &
   Location &
   Date &
  \stackbox[c]{R0\\-Value} &
  \stackbox[c]{CI \\-\% Values} &
   Method &
   Overall \\
  \hline
\multirow{4}{*}{Top 2 Highest}  & \multirow{2}{*}{s1} & Exact   & 49.28 &	47.85 &	32.82 &	46.25 &	28.30 &	27.94 &	38.77

 &
 \multirow{2}{*}{s2} & Exact   &  47.03 &	50.54 &	32.65 &	42.48 &	27.87 & 25.22 &	37.68 \\ \cline{3-10} \cline{12-19} 
                             &                             & Partial &  51.00 &	50.31 &	46.15 &	50.67 &	52.83 &	36.19 &	47.90 

 &                            & Partial & 48.58 &	52.72 &	44.90 &	50.30 &	57.38 & 35.19 &	48.31
  \\ \cline{2-19} 
                             & \multirow{2}{*}{s9} & Exact   & 47.29 &	48.02 &	31.96 &	47.21 &	26.67 &	27.67 &	38.16
 
 & \multirow{2}{*}{s1} & Exact   & 49.75 &	49.60 &	33.20 &	39.77 &	25.20 &	24.93 &	37.12
 \\ \cline{3-10} \cline{12-19} 
                             &                             & Partial &  49.00 &	50.46 &	45.36 &	51.79 &	51.43 &	37.11 & 47.58
 &                            & Partial &  50.25 &	51.19 &	45.06 &	47.13 &	55.12 &	35.13 &	47.45
  \\ \hline
\multirow{4}{*}{Top 2 Lowest} & \multirow{2}{*}{d3} & Exact   &  49.13 &	46.91 &	30.77 &	44.74 &	24.76 &	27.56 &	37.33

 & \multirow{2}{*}{d2} & Exact   &  46.80 &	48.83 &	32.13 &	39.55 &	23.26 &	24.93 &	35.94
  \\ \cline{3-10} \cline{12-19} 
                             &                             & Partial &  50.29 &	48.77 &	44.10 &	49.33 &	51.43 &	37.18 &	46.88
 &                             & Partial &  48.28 &	50.39 &	44.18 &	46.83 &	51.16 &	35.46 &	46.13

  \\ \cline{2-19} 
                             & \multirow{2}{*}{d1} & Exact   &  48.26 &	46.58 &	29.32 &	45.03 &	27.18 &	28.03 &	37.43

 & \multirow{2}{*}{d6} & Exact   &  46.42 &	48.70 &	33.33 &	39.66 &	23.44 &	25.07 &	36.13

  \\ \cline{3-10} \cline{12-19} 
                             &                             & Partial & 49.42 &	48.45 &	42.93 &	48.77 &	52.43 &	37.58 &	46.63

  &                             & Partial & 47.90 &	50.26 &	45.24 &	46.72 &	50.00 &	35.65 &	46.05

 \\ \hline
\end{tabular}
\end{adjustbox}
\caption{Top two highest and lowest inference results by \textsc{orkg}-FLAN-T5$_{R0}$ best-instructions and best-instructions with 50\% subsampled finetuned models, in descending order of overall partial F1 for the JSON answer. template column: inference instructions. SQuAD\_v2 instr. 1 (s1), SQuAD\_v2 instr. 9 (s9), DROP instr. 3 (d3), DROP instr. 1 (d1), SQuAD\_v2 instr. 2 (s2), SQuAD\_v2 instr. 1 (s1),  DROP instr. 2 (d2), and DROP instr. 6 (d6).}
\label{best-instruct-json}
\end{table*}

\section{ROUGE Evaluation Metrics}
\label{app:rouge}

The ROUGE metrics~\cite{rouge} are commonly used for evaluating the quality of text summarization systems. ROUGE-1 measures the overlap of unigram (single word) units between the generated summary and the reference summary. ROUGE-2 extends this to measure the overlap of bigram (two consecutive word) units. ROUGE-L calculates the longest common subsequence between the generated and reference summaries, which takes into account the order of words. ROUGE-LSum is an extension of ROUGE-L that considers multiple reference summaries by treating them as a single summary. These metrics provide a quantitative assessment of the similarity between the generated and reference summaries, helping researchers and developers evaluate and compare the effectiveness of different summarization approaches. They have become widely used benchmarks in the field of automatic summarization.

\section{Additional Results}
\label{app:res}

Finally, in this last appendix section, we show the highest and lowest results obtained from the two other experimental settings discussed in the main paper. I.e. all-instruction model finetuning, in two subsettings: with all the training data and with a 50\% random subsample of the training data. These results are presented in \autoref{all-instruct-text} and \autoref{all-instruct-json}, respectively, for the text format and JSON format responses. And furthermore, results are shown for the best-instruction finetuning setting in two subsettings: with all the training data and with a 50\% random subsample of the training data. These results are presented in \autoref{best-instruct-text} and \autoref{best-instruct-json}, respectively, for the text format and JSON format responses.

\end{document}